\documentclass[10pt,twocolumn,letterpaper]{article}

\usepackage[pagenumbers]{iccv} 

\newcommand{\gr}[1]{{\textcolor{gray}{#1}}}

\newcommand{\myparagraph}[1]{\smallskip \noindent \textbf{#1}}

\newcommand*\justify{%
  \fontdimen2\font=0.4em
  \fontdimen3\font=0.2em
  \fontdimen4\font=0.1em
  \fontdimen7\font=0.1em
  \hyphenchar\font=`\-
}

\def\eg{\emph{e.g}\onedot} 
\def\ie{\emph{i.e}\onedot} 
 
 \def\vs{\emph{vs}\onedot}

\def\name{ROSE}

\definecolor{cvprblue}{rgb}{0.21,0.49,0.74}
\usepackage[pagebackref,breaklinks,colorlinks,allcolors=cvprblue]{hyperref}

\definecolor{iccvblue}{rgb}{0.21,0.49,0.74}

\usepackage{framed}
\usepackage{utfsym}
\usepackage{multirow}
\usepackage{tabularx}
\usepackage{colortbl}
\usepackage{pifont}
\usepackage{multicol}
\usepackage{xcolor}
\usepackage{arydshln}
\usepackage{booktabs}
\usepackage[symbol]{footmisc}
\usepackage{amsmath}
\usepackage{amssymb}

\usepackage[linesnumbered,ruled,vlined]{algorithm2e}
\usepackage{wrapfig}

\SetKwComment{Comment}{\color{green!50!black}\# }{}

\SetKwProg{Function}{def}{:}{}

\SetKwProg{For}{for}{:}{}
\SetKwIF{If}{ElseIf}{Else}{if}{:}{elif}{else}{}%

\usepackage{etoolbox}
\makeatletter
\AfterEndEnvironment{algorithm}{\let\@algcomment\relax}
\AtEndEnvironment{algorithm}{\kern2pt\hrule\relax\vskip3pt\@algcomment}
\let\@algcomment\relax
\newcommand\algcomment[1]{\def\@algcomment{\footnotesize#1}}
\makeatother

\definecolor{LGray}{gray}{0.97}

\title{ROSE: Revolutionizing Open-Set Dense Segmentation \\
with Patch-Wise Perceptual Large Multimodal Model}

\author{Kunyang Han$^{1*}$
\hspace{0.6cm}
Yibo Hu$^{2}$\hspace{0.6cm}
Mengxue Qu$^{1*}$\hspace{0.6cm}
Hailin Shi$^{2}$\hspace{0.6cm}
Yao Zhao$^{1}$\hspace{0.6cm}
Yunchao Wei$^{1}$\\~\\
$^{1}$Beijing Jiaotong University~~~
$^{2}$NIO~~~
}

\begin{document}
\maketitle
\def\thefootnote{*}\footnotetext{Work done during internships at NIO.}

\begin{abstract}
Advances in CLIP and large multimodal models (LMMs) have enabled open-vocabulary and free-text segmentation, yet existing models still require predefined category prompts, limiting free-form category self-generation. Most segmentation LMMs also remain confined to sparse predictions, restricting their applicability in open-set environments.
In contrast, we propose ROSE, a \textbf{R}evolutionary \textbf{O}pen-set dense \textbf{SE}gmentation LMM, which enables dense mask prediction and open-category generation through patch-wise perception. 
Our method treats each image patch as an independent region of interest candidate, enabling the model to predict both dense and sparse masks simultaneously.
Additionally, a newly designed instruction-response paradigm takes full advantage of the generation and generalization capabilities of LMMs, achieving category prediction independent of closed-set constraints or predefined categories. 
To further enhance mask detail and category precision, we introduce a conversation-based refinement paradigm, integrating the prediction result from previous step with textual prompt for revision. Extensive experiments demonstrate that \name\ achieves competitive performance across various segmentation tasks in a unified framework.
Code will be released.

\end{abstract}    
\vspace{-5mm}
\section{Introduction}
\label{sec:intro}

\begin{figure}[t]
  \centering
    \includegraphics[width=1.00\linewidth]{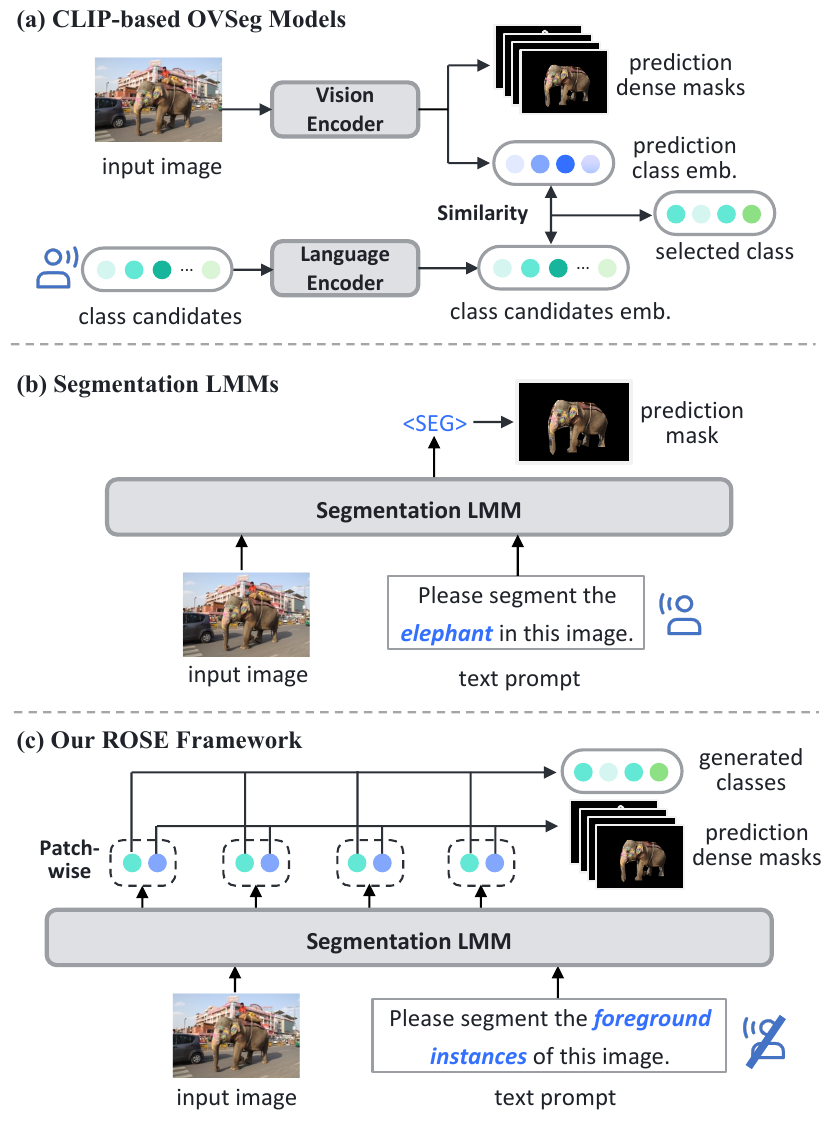}
    \caption{Comparison of existing open-set segmentation frameworks. Both (a) and (b) require predefined category inputs, where (a) uses similarity matching to select the target category, while (b) generates object masks according to the given category. Consequently, method (a) can perform dense prediction, while (b) is restricted in referring segmentation. Our approach, however, eliminates the need for predefined category inputs and produces dense predictions directly. `emb': embedding.}
    \vspace{-6mm}
  \label{fig:intro}
\end{figure}

Image segmentation is a fundamental task in computer vision, requiring pixel-level understanding and classification of image content. It reflects the fine-grained perceptual capabilities of vision models, which are crucial for accurate object recognition, scene understanding, and autonomous decision-making. Traditional segmentation methods~\cite{fcn,unet,deeplab,segformer,segnext} typically rely on fixed, closed training datasets, limiting their ability to recognize novel or unseen objects and restricting their applicability in real-world scenarios. Recent advancements in visual-language models, such as CLIP~\cite{clip}, facilitate the development of open-vocabulary segmentation (OVSeg) methods, which is able to extend the category range of traditional methods. However, these methods still depend on predefined category candidates to determine the objects to segment (Fig.~\ref{fig:intro}a), positioning them more as ``selectors" than true ``generators."

Recently, large language models (LLMs), such as LLaMA~\cite{touvron2023llama}, ChatGPT~\cite{chatgpt}, and GPT-4~\cite{openai2023gpt4}, demonstrate powerful capabilities for language understanding, reasoning, and interaction ~\cite{touvron2023llama,chatgpt,openai2023gpt4,vicuna2023,zhang2023controlvideo}, driving the emergence of large multimodal models (LMMs) like LLaVA~\cite{liu2023llava}, PaligeMMA~\cite{beyer2024paligemma}, and Ferret ~\cite{you2023ferret}. These models integrate visual and linguistic components, offering more flexibility in visual tasks. However, despite their ability to handle flexible inputs, \eg text descriptions, segmentation LMMs still rely on category prompts to determine segmentation targets (Fig.~\ref{fig:intro}b) and cannot truly ``self-generate" free-form categories. The dependence on predefined categories limits the practical applicability of existing CLIP-based open-vocabulary segmentation models and segmentation LMMs for truly open-set scenarios. 
Thus \textit{how to achieve open-set segmentation without requiring predefined category inputs} is a major challenge.

Furthermore, another key limitation with current segmentation LMMs is that most of them adopt sparse prediction rather than dense prediction, where only target regions or key objects in the image are segmented.
However, dense prediction is also crucial, with broad applications in fields such as medical image analysis and autonomous vehicle perception. It reflects the model's ability to handle complex object relationships and co-occurrence within images.
An intuitive way to adapt a Segmentation LMM (Fig.~\ref{fig:intro}b) for dense prediction is simply stacking \texttt{<SEG>} tokens in the response. However, this naive approach may result in unstable mask generation, particularly when the number of stacked \texttt{<SEG>} tokens increases. As the sequence length increases, stacked \texttt{<SEG>} tokens absorb long-range spatial dependency, which may cause the model to lose focus on local image details that are crucial for accurate segmentation. 
Therefore, \textit{how to achieve stable and efficient dense prediction while preserve fine-grained image details is another crucial challenge}.

To address these challenges, we develop a truly ``open" Segmentation LMM, termed as \name. It eliminates the requirement for predefined category inputs, and directly performs dense segmentation predictions, as illustrated in Fig.~\ref{fig:intro}c. 
Specifically, to avoid the long-range spatial dependency caused by stacked \texttt{<SEG>} tokens, we propose the Patch-wise Perception Process, which treats each image patch as an independent region of interest (RoI). Through this process, we obtain three components, including objectness score, mask embedding, and category embedding, for subsequent dense mask prediction and open-category generation. The objectness score serves to filter patches, retaining only those with high scores. We then leverage SAM to decode the mask embeddings of the selected patches, which enables our model to achieve dense mask prediction. 
Additionally, with the category embedding, we developed an instruction-response paradigm that leverages the generative and generalizable capabilities of LLMs to produce open-set category predictions, eliminating dependence on predefined category sets. In this way, the model is able to generate category names in a language-driven way, free from closed-set limitations, thus empowering the model to classify previously unseen objects autonomously.

Moreover, we introduce a conversation-based refinement mechanism to further improve segmentation details and accuracy. This paradigm allows the model to iteratively refine segmentation boundaries and categories based on user-provided text prompts, thereby enhancing the precision of segmentation masks and the accuracy of category identification, especially in complex or ambiguous visual scenes.
In extensive experiments, \name\ achieves competitive performance across various segmentation benchmarks, demonstrating its effectiveness and flexibility as an open-set dense segmentation solution. 
\name\ sets a new direction for future open-set segmentation in diverse and dynamic environments.
In summary, our contributions are as follows:

\begin{itemize}[leftmargin=4mm]
\item We present ROSE, an innovative Segmentation LMM framework that pioneers the use of patch-wise perception, enabling LMM to perform both dense and sparse mask predictions for the first time. 
\item We propose an instruction-response paradigm, fully exploiting the generative and generalizable capabilities of LLMs to achieve open-category generation.
\item With extensive experiments, we demonstrate the effectiveness of ROSE across various segmentation tasks. Additionally, a conversation-based refinement mechanism is introduced that can iteratively improve the accuracy of segmentation boundary and category prediction, particularly in complex or ambiguous visual scenes.
\end{itemize}

\section{Related works}
\label{sec:related}


\begin{table*}[th!]
    \centering
    \resizebox{0.98\textwidth}{!}{%
    \begin{tabular}{lcccccccc}
        \toprule
    \multirow{2}{*}{Method} & \multicolumn{3}{c}{Input} & \multicolumn{3}{c}{Linguistic Prompt} & \multirow{2}{*}{Dense} & \multirow{2}{*}{Segmentation} \\
    \cmidrule(lr){2-4} \cmidrule(lr){5-7} 
    & Image & Language & Region & Class Cands. & Ref. Desc. & Task Desc. & Prediction & Refinement \\
    
    \midrule
    \rowcolor{LGray}
    Mask2former (CVPR-22)~\cite{cheng2021mask2former} 
    & \textcolor{ForestGreen}{\usym{2713}} & \textcolor{red}{\usym{2717}} & \textcolor{red}{\usym{2717}} 
    & \textcolor{red}{\usym{2717}} & \textcolor{red}{\usym{2717}} & \textcolor{red}{\usym{2717}} 
    & \textcolor{ForestGreen}{\usym{2713}}  & \textcolor{red}{\usym{2717}} \\
    
    \cdashline{1-9}[1.5pt/4pt]
    OpenSeg (ECCV-22)~\cite{ghiasi2022scaling} 
    & \textcolor{ForestGreen}{\usym{2713}} & \textcolor{ForestGreen}{\usym{2713}} & \textcolor{red}{\usym{2717}} 
    & \textcolor{ForestGreen}{\usym{2713}} & \textcolor{red}{\usym{2717}} & \textcolor{red}{\usym{2717}} 
    & \textcolor{ForestGreen}{\usym{2713}}  & \textcolor{red}{\usym{2717}} \\

    \rowcolor{LGray}
    FC-CLIP (NeurIPS-23)~\cite{fcclip} 
    & \textcolor{ForestGreen}{\usym{2713}} & \textcolor{ForestGreen}{\usym{2713}} & \textcolor{red}{\usym{2717}} 
    & \textcolor{ForestGreen}{\usym{2713}} & \textcolor{red}{\usym{2717}} & \textcolor{red}{\usym{2717}} 
    & \textcolor{ForestGreen}{\usym{2713}}  & \textcolor{red}{\usym{2717}} \\
    
    \cdashline{1-9}[1.5pt/4pt]
    CascadePSP (CVPR-20)~\cite{cheng2020cascadepsp} 
    & \textcolor{ForestGreen}{\usym{2713}} & \textcolor{red}{\usym{2717}} & \textcolor{ForestGreen}{\usym{2713}} 
    & \textcolor{red}{\usym{2717}} & \textcolor{red}{\usym{2717}} & \textcolor{red}{\usym{2717}} 
    & \textcolor{red}{\usym{2717}}  & \textcolor{ForestGreen}{\usym{2713}} \\

    \rowcolor{LGray}
    SegRefiner (NeurIPS-23)~\cite{SegRefiner} 
    & \textcolor{ForestGreen}{\usym{2713}} & \textcolor{red}{\usym{2717}} & \textcolor{ForestGreen}{\usym{2713}} 
    & \textcolor{red}{\usym{2717}} & \textcolor{red}{\usym{2717}} & \textcolor{red}{\usym{2717}} 
    & \textcolor{red}{\usym{2717}}  & \textcolor{ForestGreen}{\usym{2713}} \\

    \cdashline{1-9}[1.5pt/4pt]
    LLaVA (NeurIPS-23)~\cite{liu2023llava} 
    & \textcolor{ForestGreen}{\usym{2713}} & \textcolor{ForestGreen}{\usym{2713}} & \textcolor{red}{\usym{2717}} 
    & \textcolor{red}{\usym{2717}} & \textcolor{red}{\usym{2717}} & \textcolor{red}{\usym{2717}} 
    & \textcolor{red}{\usym{2717}}  & \textcolor{red}{\usym{2717}} \\

    \rowcolor{LGray} 
    Shikra (arXiv-23)~\cite{chen2023shikra} 
    & \textcolor{ForestGreen}{\usym{2713}} & \textcolor{ForestGreen}{\usym{2713}} & \textcolor{ForestGreen}{\usym{2713}} 
    & \textcolor{red}{\usym{2717}} & \textcolor{red}{\usym{2717}} & \textcolor{red}{\usym{2717}} 
    & \textcolor{red}{\usym{2717}}  & \textcolor{red}{\usym{2717}} \\

    Kosmos-2 (arXiv-23)~\cite{peng2023kosmos} 
    & \textcolor{ForestGreen}{\usym{2713}} & \textcolor{ForestGreen}{\usym{2713}} & \textcolor{ForestGreen}{\usym{2713}} 
    & \textcolor{red}{\usym{2717}} & \textcolor{red}{\usym{2717}} & \textcolor{red}{\usym{2717}} 
    & \textcolor{red}{\usym{2717}}  & \textcolor{red}{\usym{2717}} \\

    \rowcolor{LGray} LISA (CVPR-24)~\cite{lai2024lisa} 
    & \textcolor{ForestGreen}{\usym{2713}} & \textcolor{ForestGreen}{\usym{2713}} & \textcolor{red}{\usym{2717}}
    & \textcolor{red}{\usym{2717}} & \textcolor{ForestGreen}{\usym{2713}} & \textcolor{red}{\usym{2717}} 
    & \textcolor{red}{\usym{2717}}  & \textcolor{red}{\usym{2717}} \\

    GLaMM (CVPR-24)~\cite{hanoona2023GLaMM} 
    & \textcolor{ForestGreen}{\usym{2713}} & \textcolor{ForestGreen}{\usym{2713}} & \textcolor{ForestGreen}{\usym{2713}} 
    & \textcolor{red}{\usym{2717}} & \textcolor{ForestGreen}{\usym{2713}} & \textcolor{red}{\usym{2717}} 
    & \textcolor{red}{\usym{2717}}  & \textcolor{red}{\usym{2717}} \\

    \rowcolor{LGray} 
    AnyRef (CVPR-24)~\cite{He_2024_CVPR} 
    & \textcolor{ForestGreen}{\usym{2713}} & \textcolor{ForestGreen}{\usym{2713}} & \textcolor{ForestGreen}{\usym{2713}} 
    & \textcolor{red}{\usym{2717}} & \textcolor{ForestGreen}{\usym{2713}} & \textcolor{red}{\usym{2717}} 
    & \textcolor{red}{\usym{2717}}  & \textcolor{red}{\usym{2717}} \\

    PixelLM (CVPR-24)~\cite{ren2024pixellm} 
    & \textcolor{ForestGreen}{\usym{2713}} & \textcolor{ForestGreen}{\usym{2713}} & \textcolor{red}{\usym{2717}} 
    & \textcolor{red}{\usym{2717}} & \textcolor{ForestGreen}{\usym{2713}} & \textcolor{red}{\usym{2717}} 
    & \textcolor{red}{\usym{2717}}  & \textcolor{red}{\usym{2717}} \\

    \rowcolor{LGray} 
    GSVA (CVPR-24)~\cite{xia2024gsva} 
    & \textcolor{ForestGreen}{\usym{2713}} & \textcolor{ForestGreen}{\usym{2713}} & \textcolor{red}{\usym{2717}} 
    & \textcolor{red}{\usym{2717}} & \textcolor{ForestGreen}{\usym{2713}} & \textcolor{red}{\usym{2717}} 
    & \textcolor{red}{\usym{2717}}  & \textcolor{red}{\usym{2717}} \\

    VisionLLM (NeurIPS-23)~\cite{2023visionllm} 
    & \textcolor{ForestGreen}{\usym{2713}} & \textcolor{ForestGreen}{\usym{2713}} & \textcolor{red}{\usym{2717}} 
    & \textcolor{red}{\usym{2717}} & \textcolor{ForestGreen}{\usym{2713}} & \textcolor{red}{\usym{2717}} 
    & \textcolor{red}{\usym{2717}}  & \textcolor{red}{\usym{2717}} \\

    \rowcolor{LGray} 
    PSALM (ECCV-24)~\cite{zhang2025psalm} 
    & \textcolor{ForestGreen}{\usym{2713}} & \textcolor{ForestGreen}{\usym{2713}} & \textcolor{red}{\usym{2717}} 
    & \textcolor{ForestGreen}{\usym{2713}} & \textcolor{ForestGreen}{\usym{2713}} & \textcolor{red}{\usym{2717}} 
    & \textcolor{ForestGreen}{\usym{2713}}  & \textcolor{red}{\usym{2717}} \\

    \rowcolor{blue!5}
    \textbf{\name~(Ours)} 
    & \textcolor{ForestGreen}{\usym{2713}} & \textcolor{ForestGreen}{\usym{2713}} & \textcolor{ForestGreen}{\usym{2713}} 
    & \textcolor{ForestGreen}{\usym{2713}} & \textcolor{ForestGreen}{\usym{2713}} & \textcolor{ForestGreen}{\usym{2713}} 
    & \textcolor{ForestGreen}{\usym{2713}}  & \textcolor{ForestGreen}{\usym{2713}} \\
        
    \bottomrule

    \end{tabular}
    }
    \vspace{-0.5em}
    \caption{\textbf{Comparison of the capabilities of classical models.} 
    \textit{Language} denotes the model accepts language modal as input. 
    \textit{Region} represents that the model can handle regional information.
    \textit{Linguistic Prompt} indicates which kind of linguistic information is acceptable. In which, \textit{Class Cands.} means class candidates provided by humans, \textit{Ref. Desc.} means instance-level referring description, and \textit{Task Desc.} means task-level description.
    \textit{Dense Prediction} denotes the ability to predict all targets of interest at once. 
    }
    \label{tab:methods_comparison_1}
\vspace{-1em}
\end{table*}

\subsection{Generic Segmentation}

\myparagraph{Semantic Segmentation} aims to classify each pixel in an image according to its category.
Early work FCN~\cite{fcn} uses Conv2D as the last layer and predicts category probabilities for each pixel. 
Subsequent studies focused on improving contextual understanding, some~\cite{deeplab,deeplabV3} introduced novel context modules, while others~\cite{fu2019dual,huang2019ccnet,wang2018non,zheng2021rethinking} explored self-attention mechanisms to capture pixel-wise dependencies.

\myparagraph{Instance Segmentation} aims at individually identifying each instance of a target object, 
earlier methods adopt segmentation and grouping techniques to get object proposals. This approach led to the development of bottom-up segmentation strategies~\cite{pont2016multiscale,ins_btm2up_0}, including graph-based methods~\cite{grundmann2010efficient,felzenszwalb2008discriminatively} and selective research algorithms~\cite{uijlings2013selective}.
Later, object proposals from Fast R-CNN~\cite{girshickICCV15fastrcnn} leverages for instance segmentation~\cite{Pinheiro2015,Pinheiro2016,Dai_2016_CVPR}.
SOLO~\cite{wang2020solo,wang2020solov2} advances instance segmentation by directly predicting object masks in each spatial grid.

\myparagraph{Referring Segmentation} is dedicated to segmenting a specific instance based on a natural language description. The basic principle is to merge as much linguistic information as possible to the visual feature~\cite{res1,res2_VLT,res5,res6}.
Previous methods focus on the various attention mechanisms~\cite{lts,res2_VLT} to better incorporate language and vision. Recently, with the success of transformer-based models on vision and language area, some powerful works\cite{qu2023learning,zhu2022seqtr,li2021restr} come off.

\subsection{Vision Language Models}

The emergence of LLMs~\cite{touvron2023llama,chatgpt,openai2023gpt4,vicuna2023} has led to notable developments in vision-language modeling, where models are designed to understand both visual and textual inputs. Foundational works~\cite{li2023blip,dai2023instructblip,zhu2023minigpt,liu2023llava} focus on aligning visual features with language representations, although they are limited in their applicability to region-level tasks.

Recently, vision-language models such as Kosmos~\cite{kosmos-2} and All-Seeing~\cite{wang2023allseeing} achieved region grounding by employing bounding box-based formats, while models like GPT4RoI~\cite{zhang2023gpt4roi} and Ferret~\cite{you2023ferret} introduced region-based encoders for enhanced understanding of visual regions.
Furthermore, LISA~\cite{lai2024lisa} introduces the \texttt{<SEG>} token for pixel-level referring segmentation, with subsequent models like PixelLM~\cite{ren2024pixellm} and GSVA~\cite{xia2024gsva} extending this approach to multi-target referring segmentation.
GLaMM~\cite{hanoona2023GLaMM} introduced a hierarchical feature pyramid for regional prompting, while 
CoRes~\cite{bao2024cores} incorporated a CoT procedure to improve contextual understanding in segmentation.
VisionLLM~\cite{2023visionllm} develops a set of prompts to handle various visual segmentation tasks.

\subsection{Open-set Image Segmentation}
Recent methods\cite{ghiasi2022scaling,zegformer,zsseg} are built on the MaskFormer framework~\cite{cheng2021maskformer}, they generate class-agnostic masks and compare the similarity with text embeddings from models like CLIP~\cite{radford2021learning} and ALIGN~\cite{jia2021scaling} to classify these regions.
OpenSeg~\cite{ghiasi2022scaling} utilizes image-level supervision and scales the training data, while models like Zegformer~\cite{zegformer} and ZSSeg~\cite{zsseg} improved precision by cropping and refining sub-images before processing them with CLIP.
Based on them, GKC~\cite{han2023global} enhances vision-text alignment by generating synonyms. OVSeg~\cite{ovseg} trains a CLIP adapter to boost the performance. ODISE~\cite{odise} introduces a strong text-to-image diffusion model to learn the text feature space. FC-CLIP~\cite{fcclip} designs an end-to-end framework that uses a single frozen CLIP as the backbone. MAFT+~\cite{jiao2024collaborative} proposes a collaborative framework to optimize vision-text representation jointly.
Recently, PSALM~\cite{zhang2025psalm} replaced the CLIP vision encoder and the transformer decoder of MaskFormer with a large language model~\cite{li2023textbooksneediiphi15}.

\begin{figure*}[t]
  \centering
    \includegraphics[width=0.98\linewidth]{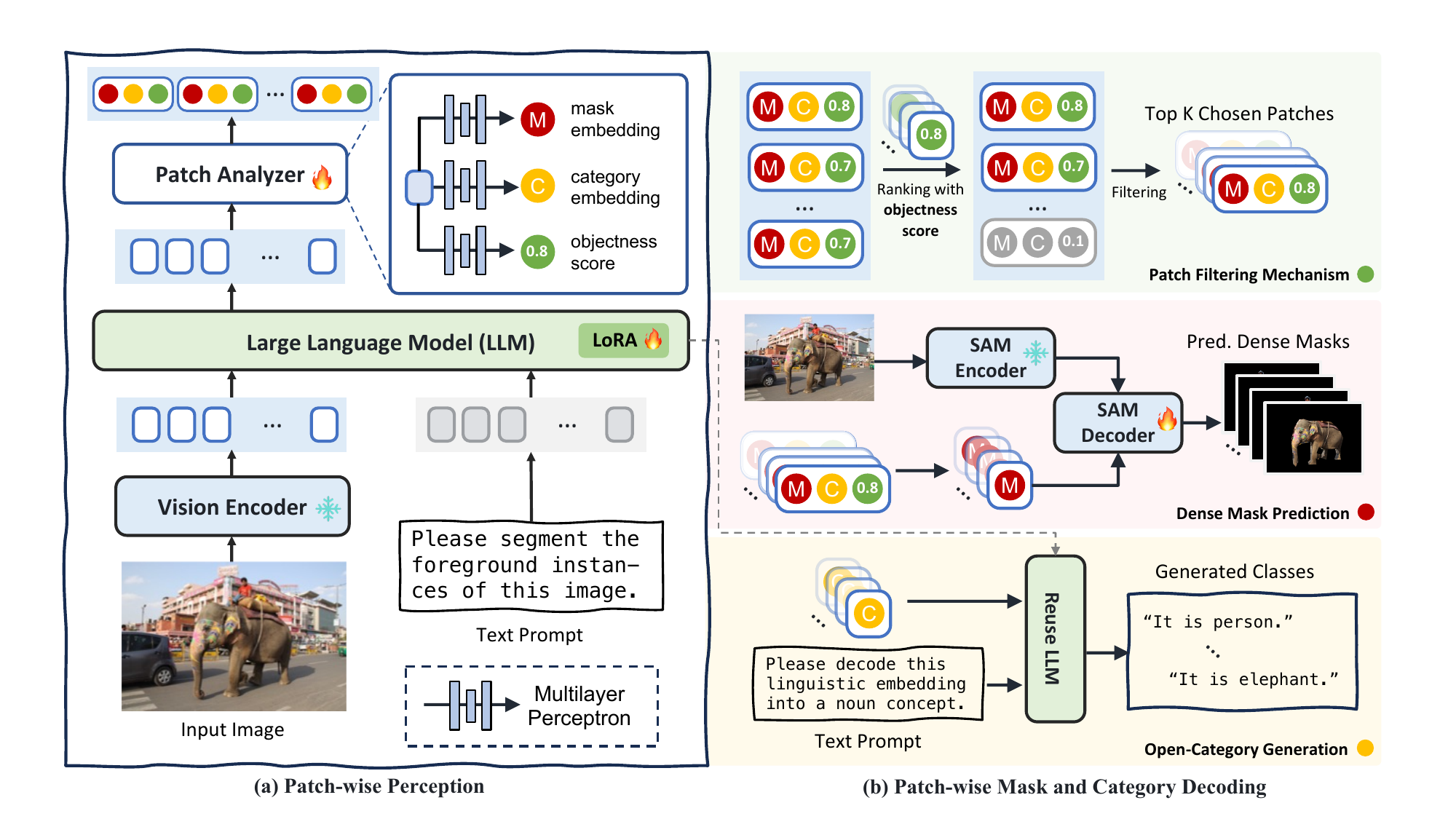}\vspace{-0.5em}
    \caption{\textbf{The architecture of \name.} (a) \textit{In Patch-wise Perception Processes}, the vision encoder first encodes the input image and gets patched features, the feature is then concatenated with text instruction and fed into the Large Language model. Then every patch is analyzed by the patch analyzer, generating a mask embedding, a category embedding, and an objectness score. (b) \textit{In Patch-wise Mask and Category Decoding Process}, patches are first filtered with objectness scores. Then mask embedding is fed into the SAM decoder as a prompt for the patch-corresponding mask. Category embedding is employed to make corresponding category predictions in a generative way.}
    \label{fig:pipiline}
\vspace{-1em}
\end{figure*}

\section{Method}

In this section, we first define the task in Sec.~\ref{sec: task_define}, and then detail our \name\ framework in Sec.~\ref{sec: patch_wise_perception} - Sec.~\ref{sec: refinement}. Finally, we outline the training objectives in Sec.~\ref{sec: objectives}.

\subsection{Task Definition}
\label{sec: task_define}

Given an image $\mathbf{x}_\text{img}$ and a text instruction $\mathbf{x}_\text{txt}$, the goal is to complete the segmentation procedure, generating a segmentation mask $\hat{\mathbf{M}}$ and category $\hat{\mathbf{y}}_\text{txt}$. \textbf{Open-vocabulary methods} require $\mathbf{x}_\text{txt}$ to consist of a set of human-provided candidate categories. These methods predict a list of masks $\hat{\mathbf{M}} \in \mathbb{R}^{N \times H \times W}$ and select $\hat{\mathbf{y}}_\text{txt}$ from the candidate set. Existing \textbf{Segmentation LMMs}, on the other hand, typically require $\hat{\mathbf{y}}_\text{txt}$ to be an instance-level description, which allows them to predict the corresponding target mask $\hat{\mathbf{M}} \in \mathbb{R}^{H \times W}$ without category information.

Our goal is to propose a Segmentation LMM that can not only segment objects based on human prompts or predefined categories but also autonomously predict dense segmentation masks without any human-provided information. 
We refer this task as \textbf{Free-vocabulary Segmentation}, 
where the model accepts task-level instructions (e.g., “Can you segment the foreground instance?”), and generates a dense segmentation $\hat{\mathbf{M}} \in \mathbb{R}^{N \times H \times W}$ along with the predicted category $\hat{\mathbf{y}}_\text{txt}$ in a generative manner.

\subsection{Patch-wise Perception} 
\label{sec: patch_wise_perception}

Embedding-based mask generation, as proposed by LISA~\cite{lai2024lisa}, offers a solution to mask prediction. However, directly stacking \texttt{<SEG>} tokens proves inadequate for dense object prediction (shown in Tab.~\ref{tab:seg_framework}). 
Contemporary LMMs typically rely on ViT-based encoders or raw image patches for feature processing, handling image data in a patch-wise manner. Inspired by the SOLO instance segmentation model~\cite{wang2020solo}, which divides images into grid-based predictions, our model takes image patches as fundamental prediction units, enabling object detection on a finer scale. Our patch-wise perception process is shown in Fig.~\ref{fig:pipiline}a. 
Specifically, each patch predicts the following three components: 
1) objectness score, indicates the probability of an object of interest being present within the current patch. 2) Mask embedding, serves as input to the SAM module for mask generation. 3) Category embedding, utilized for subsequent classification tasks.
To implement this, we begin by dividing an input image $\mathbf{x}_\text{img}$ of size $L$$\times$$L$ into non-overlapping patches of size $p$$\times$$p$, yielding $S$$\times$$S$ patches, where $S=\lfloor \frac{L}{p} \rfloor$. Thus, the maximum number of predictions is $S^2$.
If a target object’s mass center falls within the spatial region of a patch located at coordinates $(h,w)$, supervision is assigned to that patch and all 8 patches surround it $(h\pm1,w\pm1)$.
All the $S^2$ patches are first passed into vision encoder $\mathcal{F}_\text{vis}$, and the resulting visual features, combined with the task instruction $\mathbf{x}_\text{txt}$, are further processed by LLM $\mathcal{F}_\text{llm}$:

\vspace{-10pt}
\begin{align}
\begin{aligned}
    \hat{\mathbf{y}}_\text{txt} = \mathcal{F}_\text{llm}(\mathcal{F}_\text{vis}(\mathbf{x}_\text{img}), \mathbf{x}_\text{txt}).
\end{aligned}
\end{align}

We also proposed a mechanism called super-patch, which clusters patches from nearby, designates specialized detecting roles based on object scale (\ie, small, medium, large) and type (thing \vs stuff categories). This role-based segmentation allows the model to adapt predictions based on object characteristics, enhancing performance across varied visual tasks and object scales.

For patches identified as containing target objects, object embeddings $\mathbf{E}_\text{obj}$ are extracted from the last layer of LLM. These embeddings are fed into distinct Multilayer Perceptrons (MLPs) to predict the objectness score, mask embedding $\mathbf{E}_\text{msk}$, and category embedding $\mathbf{E}_\text{cat}$:

\vspace{-10pt}
\begin{align}
\begin{aligned}
    \text{objectness} = & \; \phi_\text{obj}(\mathbf{E}_\text{obj}), \\
    \mathbf{E}_\text{msk} = \phi_\text{msk}(\mathbf{E}_\text{obj}),& \; \mathbf{E}_\text{cat} = \phi_\text{cat}(\mathbf{E}_\text{obj}).
\end{aligned}
\end{align}

Beyond the aforementioned three components, we also assign more MLPs to predict the SigLIP~\cite{zhai2023sigmoid} embedding, which aims to align with the latent space of the SigLIP~\cite{zhai2023sigmoid} text encoder.

\begin{table*}[th!]
    \centering
    \resizebox{0.98\textwidth}{!}{%
    \begin{tabular}{lcccccccccc}
    \toprule

    \multirow{3}{*}{Method} & Semantic Seg & Instance Seg & \multicolumn{8}{c}{Referring Segmentation} \\
    & \multirow{2}{*}{ADE-20k} & \multirow{2}{*}{COCO} & \multicolumn{3}{c}{RefCOCO} & \multicolumn{3}{c}{RefCOCO+} & \multicolumn{2}{c}{RefCOCOg} \\
    \cmidrule(lr){4-6} \cmidrule(lr){7-9} \cmidrule(lr){10-11}
    & & & val & testA & testB & val & testA & testB & val(U) & test(U) \\
    
    \midrule
    
    \textit{\small specialist model} &  &  &  &  &  &  &  &  &  &  \\
    \gr{Mask2former~\cite{cheng2021mask2former}} & \gr{57.7} & \gr{50.1} & \gr{-} & \gr{-} & \gr{-} & \gr{-} & \gr{-} & \gr{-} & \gr{-} & \gr{-} \\
    
    \arrayrulecolor{black}
    \cdashline{1-11}[1.5pt/4pt]
    \textit{\small generalist model} &  &  &  &  &  &  &  &  &  &  \\
    Painter~\cite{Painter} & 43.4 & - & - & - & - & - & - & - & - & - \\
    \rowcolor{LGray} 
    SegGPT~\cite{SegGPT} & 34.4 & - & - & - & - & - & - & - & - & - \\
    Pix2Seq v2~\cite{chen2022unified} & - & \underline{38.2}$^{\dag}$ & - & - & - & - & - & - & - & - \\
    \rowcolor{LGray} 
    PSALM~\cite{zhang2025psalm} & - & - & \underline{83.6} & \underline{84.7} & \underline{81.6} & 72.9 & 75.5 & \underline{70.1} & 73.8 & 74.4 \\
    Osprey-7B~\cite{Osprey} & 29.6$^{*}$ & - & - & - & - & - & - & - & - & - \\
    \rowcolor{LGray} 
    LISA-7B~\cite{lai2024lisa} & - & - & 74.9 & 79.1 & 72.3 & 65.1 & 70.8 & 58.1 & 67.9 & 70.6 \\
    GLaMM-7B~\cite{hanoona2023GLaMM} & - & - & 79.5 & 83.2 & 76.9 & 72.6 & \underline{78.7} & 64.6 & 74.2 & 74.9 \\
    \rowcolor{LGray} 
    GSVA-7B~\cite{xia2024gsva} & - & - & 77.2 & 78.9 & 73.5 & 65.9 & 69.6 & 59.8 & 72.7 & 73.3 \\
    GSVA-13B~\cite{xia2024gsva} & - & - & 79.2 & 81.7 & 77.1 & 70.3 & 73.8 & 63.6 & 75.7 & \underline{77.0} \\
    \rowcolor{LGray} 
    Ferret-7B~\cite{you2023ferret} & 31.8$^{*}$ & - & - & - & - & - & - & - & - & - \\
    VisionLLM-7B~\cite{2023visionllm} & - & 30.6$^{\dag}$ & - & - & - & - & - & - & - & - \\

    \rowcolor{blue!5}
    \textbf{\name-7B}       & \underline{51.0} & 36.3 & 80.1 & 81.9  & 76.9 & \underline{73.1} & 78.5 & 67.3 & \underline{75.9} & 75.6  \\
    \rowcolor{blue!5}
    \textbf{\name-7B + CSR} & \textbf{57.4} & \textbf{39.1} & \textbf{87.2} & \textbf{87.8} & \textbf{86.0} & \textbf{87.0} & \textbf{87.3} & \textbf{86.0} & \textbf{85.6} & \textbf{86.1} \\

    \bottomrule

    \end{tabular}
    }
    \vspace{-0.5em}
    \caption{\textbf{Comparison with SOTA models on common generic segmentation benchmarks.} We evaluate our model on ADE-20k dataset for semantic segmentation, COCO dataset for instance segmentation, and refCOCO/+/g for referring segmentation. $^{*}$ denotes the paradigm that the model generates a regional description based on the GT mask, both of the results come from Osprey. $^{\dag}$ denotes the framework that the model predicts masks according to the provided category. The \textbf{best result} and \underline{second best result} are highlighted in bold and underlined.
    }
    \label{tab:main_exp}
\vspace{-1em}
\end{table*}

\subsection{Patch-wise Mask and Category Decoding}
\label{sec: patch_decoding}
After obtaining the objections score, mask embedding $\mathbf{E}_\text{msk}$, and category embedding $\mathbf{E}_\text{cat}$, we further leverage these components to decode the mask and category. An illustration is shown in Fig.~\ref{fig:pipiline}b with the detailed process as follows.

\noindent{\textbf{Patch Filtering Mechanism.}} The objectness score serves as a filtering mechanism during inference, allowing the model to prioritize patches with high confidence for further segmentation processing. 
In order to make $\phi_\text{obj}$ converge, we will randomly pick some unsupervised patches as negative samples of objections.

\noindent{\textbf{Dense Mask Prediction.}} 
Following LISA~\cite{lai2024lisa}, mask embedding $\mathbf{E}_\text{msk}$ is fed into the SAM decoder $\mathcal{F}_\text{dec}$ as the text prompt, which generates the final mask prediction:

\vspace{-10pt}
\begin{align}
\begin{aligned}
    \mathbf{f} = \mathcal{F}_\text{enc}(\mathbf{x}_\text{img}), 
    \hat{\mathbf{M}} = \mathcal{F}_\text{dec}(\mathbf{E}_\text{msk}, \mathbf{f}).
\end{aligned}
\end{align}

\noindent $\mathcal{F}_\text{enc}$ is the encoder of SAM, which extracts the image feature $\mathbf{f}$ from input image $\mathbf{x}_\text{img}$.

\noindent{\textbf{Open-category Generation.}} In contrast to ~\cite{lai2024lisa,2023visionllm,ren2024pixellm,xia2024gsva} concentrate on reasoning and neglect classification and ~\cite{wang2024llmseg, zhang2025psalm} adopt similarity-comparison paradigm, 
we employ a generative approach where the model produces category predictions through language generation, independent of predefined category constraints.
An intuitive adaptation is employing random or learnable queries inside each Patch Analyzer to generate categories. 
However, the number of such queries is unpredictable, caused by the uncertainty that both the number of words that make up a category and the number of tokens represent the word (\eg ``staircase'' worth three tokens for the tokenizer).
This problem could be solved by adding a sufficient number of queries, but this will bring great computing costs.

To address this issue, we treat the category embedding $\mathbf{E}_\text{cat}$ as a linguistic feature, allowing us to implement a custom instruction-response paradigm for the classification ``\texttt{\textbf{USER}}: \texttt{<CATEGORY>} \texttt{\small \justify Please decode this linguistic embedding into a noun concept.} \texttt{\textbf{ASSISTANT}}: \texttt{\small Sure, it is} \texttt{\{category\_name\}}."

\noindent
Here, \texttt{<CATEGORY>} is the category embedding $\mathbf{E}_\text{cat}$, and \textit{\{category\_name\}} is supposed to be the specific words of target category as the prediction. Specifically, if there are $N$ activated detecting patches, the shape of the input instruction $\mathbf{I}_\text{cat}$ will be $N \times L_\text{seq} \times D$, in which $L_\text{seq}$ is the length of the input sequence, and $D$ represents the hidden dimension. 

\vspace{-10pt}
\begin{align}
\begin{aligned}
    \hat{\mathbf{y}}_\text{cat} = \mathcal{F}_\text{llm}(\mathbf{E}_\text{cat}, \mathbf{I}_\text{cat}).
\end{aligned}
\end{align}

In the training stage, $N$ is determined by the number of ground truth (GT), and \textit{\{category\_name\}} $\mathbf{y}_\text{cat}$ is provided in the instruction with an attention mask. In the inference, $N$ is a configurable parameter; the detecting patches with top-$N$ objectness score are kept to produce the final prediction. And \textit{\{category\_name\}} $\hat{\mathbf{y}}_\text{cat}$ is generated in an auto-regressive manner simultaneously for all $N$ targets.

\subsection{Conversation-based Segmentation Refinement}
\label{sec: refinement}

Recent chain-of-thought works~\cite{cot1,cot2} demonstrate that if more explicit instructions are given, LLMs have the potential to sense the details and correct themselves. If LLM can correct its language wrongness, then LMM might also be able to refine its segment predictions and eventually get better results. 
Following this thought, we propose the CSR paradigm, in which the model takes in the image $\mathbf{x}_\text{img}$, mask $\hat{\mathbf{M}}$, category $\hat{\mathbf{y}}_\text{cat}$, and refinement instruction $\mathbf{I}_\text{ref}$, using these elements to refine segmentation predictions. 
If one wants an LMM to refine the result, they must let the LMM understand it.
To accomplish this, some work~\cite{zhang2024groundhog} introduced DINO~\cite{caron2021emerging} and some~\cite{you2023ferret} use pooling-based processing. However, for the simplicity and maintenance of spatial information, we choose to concatenate images and masks directly.

It is worth noting that mask $\hat{\mathbf{M}}$ and category $\hat{\mathbf{y}}_\text{cat}$ are not necessarily used according to the refinement scenarios. Overall, we define three key cases: 1) correct classification with imperfect segmentation. 2) incorrect classification with imperfect segmentation. 3) missed detections.
Categories are provided in the second case in instruction, and segment masks are used in the first two situations. A full-zero tensor will be concatenated with the image when the mask is absent.
Each case is supported by ten unique instructions, with tailored bounding box information to focus the model's attention on the target.

\subsection{Training Objectives}
\label{sec: objectives}

Our training process optimizes five objective functions:
1) Text generation loss, we use Cross-Entropy loss to supervise $\hat{\mathbf{y}}_\text{txt}$ and $\hat{\mathbf{y}}_\text{cat}$.
As mentioned before, $\mathbf{y}_\text{cat}$ is the plain category name,
and we design a counting task for $\mathbf{y}_\text{txt}$, it looks like
``\texttt{\textbf{Assistant}}: \texttt{\small There are 5 person, 2 bicycle, 4 car, 2 truck, 1 umbrella in this image.}".
2) Following LISA, our mask prediction loss is composed of Dice loss and Binary Cross-entropy (BCE) loss. With the GT mask $\mathbf{M}$ and category $\mathbf{y}_\text{cat}$, these can be formulated as:

\vspace{-10pt}
\begin{align}
\begin{aligned}
    \mathcal{L}_{txt} = \mathbf{CE}(\hat{\mathbf{y}}_{txt}, \mathbf{y}_{txt}) &+ \mathbf{CE}(\hat{\mathbf{y}}_{cat}, \mathbf{y}_{cat}), \\
    \mathcal{L}_{mask} = \lambda_{bce} \mathbf{BCE}(\hat{\mathbf{M}}, \mathbf{M})  &+ \lambda_{dice}\mathbf{DICE}(\hat{\mathbf{M}}, \mathbf{M}).
\end{aligned}
\end{align}

\noindent
3)  For objectness loss $\mathcal{L}_{obj}$ we utilize BCE loss. The GT of the patches with the center of the target around is set to positive, and other patches that are extra-picked are set to negative. 4) SigLIP embedding loss $\mathcal{L}_{sig}$ adopts InfoNCE supervision to learn the latent space of the text encoder from SigLIP. The overall loss is formulated as follows: 

\vspace{-10pt}
\begin{align}
\begin{aligned}
    \mathcal{L} = \mathcal{L}_{txt} + \mathcal{L}_{mask} + \mathcal{L}_{obj} + \mathcal{L}_{sig}
\end{aligned}
\end{align}
\section{Experiment}

\subsection{Experiment setting}

\myparagraph{Network Architecture.}
We use llava-onevision-7b~\cite{li2024llava} as LMM, which contains SigLIP~\cite{zhai2023sigmoid} $\mathcal{F}_\text{vis}$ and Qwen~\cite{qwen} $\mathcal{F}_\text{llm}$, and adopt ViT-H SAM~\cite{sam} as $\mathcal{F}_\text{enc}$ and $\mathcal{F}_\text{dec}$. The projectors $\phi$ used to generate embeddings $\mathbf{E}$ are two-layer MLP with a ReLU~\cite{relu} activation and hidden dimension of 3584.
During training, to preserve the knowledge of llava, we employ LoRA~\cite{hu2021lora} on Qwen $\mathcal{F}_\text{llm}$. SigLIP $\mathcal{F}_\text{vis}$ and SAM vision encoder $\mathcal{F}_\text{enc}$ are both frozen. New added projectors $\phi$ and SAM decoder $\mathcal{F}_\text{dec}$ are fully fine-tuned. Additionally, the head layer (\texttt{lm\_head}) and token embedding (\texttt{embed\_tokens}) of Qwen, and the patch layer (\texttt{patch\_embedding}) of SigLIP are also trainable.

\myparagraph{Implementation Details.}
Our implementation is based on deepspeed~\cite{rasley2020deepspeed}. 8 NVIDIA A800 GPUs are adopted for training.
The optimizer is AdamW~\cite{adamw} with a learning rate of 0.003. We use WarmupDecayLR as the learning rate scheduler, and a linear 1500-iteration warmup is set. The total training iteration is 50k, the per-device batch size is 2, and the gradient accumulation step is 10. The input image is resized to 672 $\times$ 672. There are $48^2$ predicting patches, and we select the top 100 patches from it for inference.

\myparagraph{Dataset and Task.}
Our training tasks and datasets are composed of the following: 1) Semantic segmentation, we use ADE20k~\cite{ade}, COCO-Stuff~\cite{cocostuff}, and Mapillary~\cite{mapillary} in this task. 2) Instance segmentation, In this task, we only use one dataset, COCO~\cite{coco}, in the training. 3) Referring segmentation, Following LISA, we use RefCLEF, RefCOCO, RefCOCO+~\cite{refcoco}, and RefCOCOg~\cite{refcocog}. 
4) Segmentation Refinement. 
We generate defective masks from GT masks to collect training pairs in two ways:
the first follows ~\cite{cheng2020cascadepsp} randomly added some holes and extra patches, and the second randomly shrinks or stretches the object area and keeps parts of this variation.

\begin{table}[t]
    \centering
    \small
    \resizebox{\linewidth}{!}{
    \begin{tabular}{lc|cc}
        \toprule
        Segment Unit & Classify Uint & ADE-20k & COCO \\ 
        \midrule
        Vanilla stack & Along \texttt{<SEG>} & 37.2 & 23.2 \\ 
        Dense stack & Along \texttt{<SEG>} & 34.4 & 22.6  \\ 
        Patch-wise & Along \texttt{<REGION>} & 21.5 & 4.4 \\
        Patch-wise & Decode embed & \textbf{47.5} & \textbf{29.2} \\
        \bottomrule
        \vspace{-15pt}
    \end{tabular}
    }
    \caption{Ablation study of different segmentation framework.}
    \vspace{-6pt}
    \label{tab:seg_framework}
\end{table}

\begin{table}[t]
    \centering
    \small
    \resizebox{\linewidth}{!}{
    \begin{tabular}{l|cccc}
        \toprule
        \multirow{2}{*}{Patch-design} & \multirow{2}{*}{ADE-20k} & \multicolumn{3}{c}{COCO} \\ 
        \cmidrule(lr){3-5}
        & & mAP & AP50 & AP75 \\
        \midrule
        w/o super-patch & \textbf{47.5} & 29.2 & 45.7 & 31.4 \\ 
        $2$$\times$$2$  & 43.0 & 29.8 & 45.4 & 32.3 \\ 
        $3$$\times$$3$  & 43.2 & \textbf{32.4} & \textbf{49.2} & \textbf{34.8} \\
        \bottomrule
        \vspace{-15pt}
    \end{tabular}
    }
    \caption{Ablation study of different super-patch.}
    \vspace{-8pt}
    \label{tab:super_patch}
\end{table}

\subsection{Generic Segmentation}
To evaluate the effectiveness of our proposed \name, we conduct experiments to demonstrate its capabilities on 
three common segmentation tasks.
The prediction of original ROSE and ROSE with conversation-based segmentation refinement (CSR) is reported in Tab.~\ref{tab:main_exp}.

\myparagraph{Semantic Segmentation.} 
In this task, Painter and SegGPT achieving 43.4 and 34.4 mIoU correspondingly. 
Osprey and Ferret adopt ground-truth masks to get sentence-based responses and calculate their similarity to the vocabulary list to get the category predictions. They score 29.6 and 31.8 mIoU correspondingly.
The direct prediction from \name\ gets the result of 43.2 mIoU, slightly lower than Painter. However, CSR largely boosts performance to 51.6 mIoU and achieves SOTA.

\myparagraph{Instance Segmentation.} This task is more difficult than semantic segmentation because it requires identity information and confidence score in addition. 
Pix2Seq v2 and VisionLLM adopt the GT category in the prompt and yield class-agnostic masks, they achieve 38.2 and 30.6 mAP correspondingly.
\name\ takes the objectness score from the patch analyzer as confidence score, and has a comparable performance of 34.4 mAP.

\myparagraph{Referring Segmentation.} This task requires the model to understand linguistic instruction, LMMs that use LMMs have natural advantages on it. Ferret concentrates on regional understanding and performs greatly. Our model shows a comparative referring ability in competition with a larger model (Ferret-13B) and achieves multiple SOTAs in refcoco/+/g datasets.

\myparagraph{Conversation-based Segmentation Refinement.} This task largely unleashes the potential of LMM and boosts its performance. In the semantic segmentation task, 
we refine the five worst categories predictions, which bring 8.6 mIoU boosts. 
In instance segmentation, we pick up 10 under-IoU-threshold predictions in the descending order of objectness score and bring 2.0 mAP improvements.
In referring segmentation, CSR brings 12.6 $\pm$ 4.4 gains on average.

\begin{table}[t]
    \centering
    \small
    \resizebox{\linewidth}{!}{
    \begin{tabular}{lc|cc}
        \toprule[1pt]
        Target Modules & LoRA Alpha & ADE-20k & COCO \\ 
        \midrule
        \texttt{q\_proj,k\_proj}  & 16 & 24.7 & 19.2 \\ 
        \texttt{q\_proj,k\_proj}  & 32 & 25.1 & 19.5 \\ 
        \texttt{all\_proj}       & 16 & 26.3 & 22.7 \\ 
        \texttt{all\_proj}       & 32 & \textbf{29.1} & \textbf{23.1} \\ 
        \bottomrule[1pt]
        \vspace{-15pt}
    \end{tabular}
    }
    \caption{Ablation study of different LoRA parameters.}
    \vspace{-9pt}
    \label{tab:abl_lora}
\end{table}
\begin{table}[t]
    \centering
    \small
    \resizebox{\linewidth}{!}{
    \begin{tabular}{l|cccc}
        \toprule
        \multirow{2}{*}{Method} & \multirow{2}{*}{ADE-20k} & \multicolumn{3}{c}{COCO} \\ 
        \cmidrule(lr){3-5}
        & & mAP & AP50 & AP75 \\
        \midrule
        Mask           & $\uparrow$ \textbf{6.4} & $\uparrow$ 3.0 &  $\uparrow$ 4.0 & $\uparrow$ 3.1 \\ 
        Mask + Bbox    & $\uparrow$ 6.3 & $\uparrow$ 3.0 &  $\uparrow$ 5.2 & $\uparrow$ 2.8 \\ 
        Concatenation  & $\uparrow$ 5.4 & $\uparrow$ \textbf{3.6} &  $\uparrow$ \textbf{6.4} & $\uparrow$ \textbf{3.7} \\
        \bottomrule
        \vspace{-15pt}
    \end{tabular}
    }
    \caption{Ablation study of different refinement paradigms.}
    \vspace{-10pt}
    \label{tab:refine}
\end{table}

\subsection{Ablation Study}
\label{sec:ablation}

\myparagraph{Segmentation Framework.}
We explore the effectiveness of different paradigms for dense prediction frameworks, results are shown in the Tab.~\ref{tab:seg_framework}. Vanilla stack means stacking $N$ \texttt{<SEG>} together for $N$ targets. Along \texttt{<SEG>} means generate categories ahead of \texttt{<SEG>}. Our experiment shows it performs badly.
Dense stack also performs \texttt{<SEG>} stacking but adds more \texttt{<SEG>} for each GT target. The result shows that simply adding more supervision is not helpful.
Patch-wise is used by \name, it takes the image patches as the predicting unit. 
Along \texttt{<REGION>} use \texttt{<REGION>} to indicate each patch and generate the corresponding category, our further investigation finds out it causes mismatches between mask and category.
By generating categories and decoding them, \name\ gets a more stable and excellent performance.

\myparagraph{Super-patch Design.}
Using patch-wise prediction and decoding manner classification leads \name\ to good performance. Still, it suffers from bad results on instance segmentation. 
Inspired by PixelLM~\cite{ren2024pixellm}
we propose the super-patch. With spatial $3$$\times$$3$ patches assembled, we assign 4 patches for small object detection, 3 patches for medium object, and one each for large object and stuff region. 
Results in Tab.~\ref{tab:super_patch} 
show it drags down the performance on semantic segmentation, it may caused by the insufficient number of stuff detectors. 
But with such multi-scale role-splitting, instance segmentation gets a great gain.
We also conduct a $2$$\times$$2$ super-patch design but the performance gain is limited.

\begin{figure*}[t]
  \centering
    \includegraphics[width=0.98\linewidth]{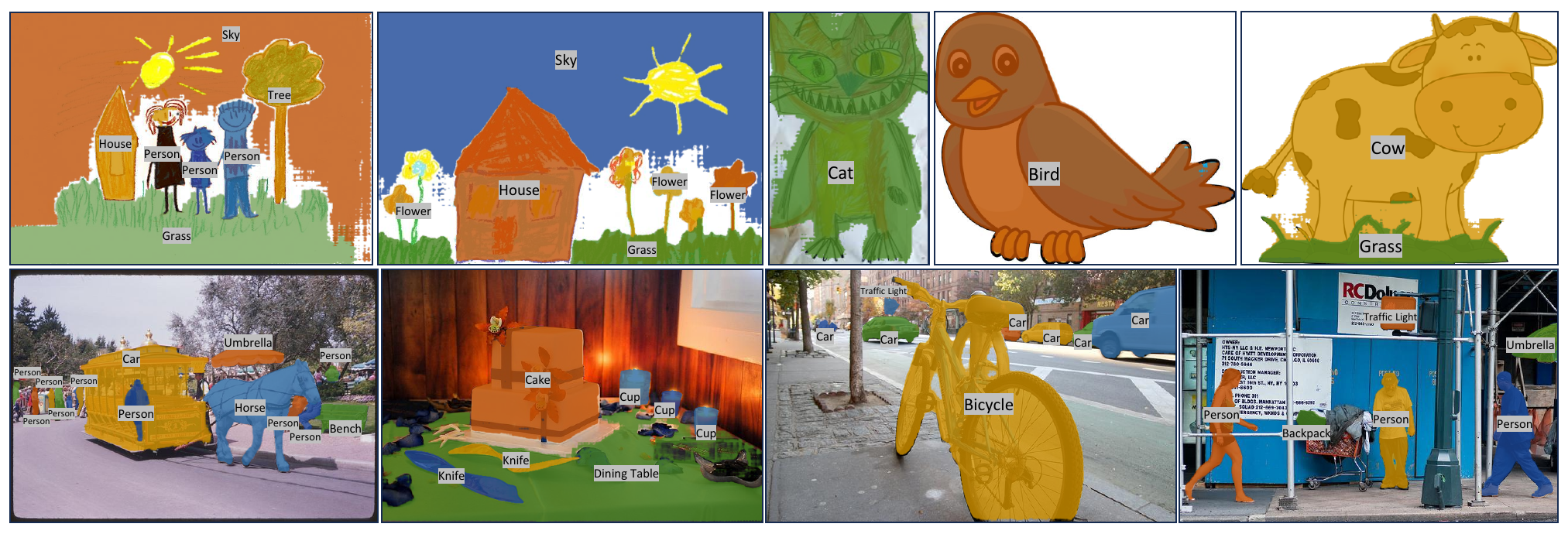}\vspace{-0.5em}
    \caption{\textbf{Qualitative results.} We show some predictions of \name\ in cross-domain and in-domain scenarios, with generated categories labeled near each target. Please zoom in to see the details. 
    The first row shows the results of images from other domains, including crayon drawings and clip art. The second row shows some predictions of the COCO val set.
    }
    \label{fig:visual}
\vspace{-1em}
\end{figure*}

\begin{figure*}[t]
  \centering
    \includegraphics[width=0.98\linewidth]{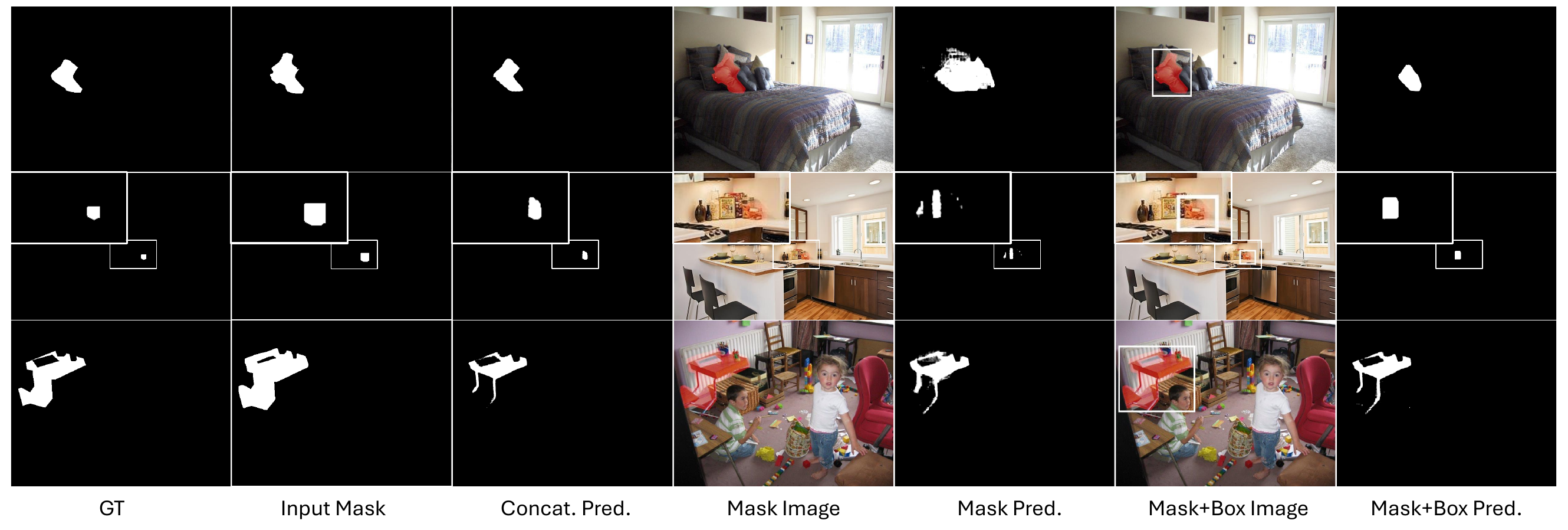}\vspace{-0.5em}
    \caption{\textbf{Visualization of different refinement mechanisms.} The first two columns are ground truth and mask expected to be refined. \textit{Concat.} denotes concatenate mask with image, and \textit{Pred.} stands for prediction. \textit{Mask} and \textit{Mask+Box} are other methods we try. 
    }
    \label{fig:refine}
\vspace{-1em}
\end{figure*}

\myparagraph{LoRA Parameter.}
We investigate the effect of the target module and LoRA alpha in Tab.~\ref{tab:abl_lora}. 
It shows as the target modules and LoRA alpha increase, the performance continues to improve.
It is worth noting that these models are trained with 20\% iterations compared to others. 

\myparagraph{Refinement Mechanism.}
To make \name\ understand and refine its past predictions, we conduct several different paradigms. The performance gain is shown in the Tab.~\ref{tab:refine}
Besides the concatenating we mentioned above, 
following FGVP~\cite{yang2023finegrainedvisualprompting}, we also run another two experiments: 1) draw the segment mask on the image, and 2) draw the segment mask and bounding box on the image. 
However, we find that both of them cause target-shifting problems occasionally (shown in Fig.~\ref{fig:refine} first two rows), especially when counting on small regions. We can tell such degradation from the COCO dataset, in which there are more small objects. 
We run the evaluation experiments on a subset with 1k samples for time efficiency.

\subsection{Qualitative Results}
\label{sec:qualitative}
As depicted in Fig.~\ref{fig:visual}, we present the predictions of ROSE in cross-domain and in-domain scenarios. Experiments demonstrate that the model can autonomously and accurately classify and segment instance objects, both within in-domain (row 2) COCO scenes and in cross-domain (row 1) crayon drawings and clip art scenes.
\section{Conclusion}
In this paper, we presented \name, a novel framework enabling dense mask prediction and open-category generation across the image. 
We designed the patch-wise perception process, which treats each image patch as an independent region, addressing the dense prediction problem. 
We also proposed a new instruction-response paradigm, allowing the model to classify in a generative way. 
To further unleash the power of LMM, we introduced a conversation-based refinement mechanism, which largely boosts the performance.
We hope our work shows a creative perspective for the coming open-set segmentation works.

\myparagraph{Limitation} Our approach advances open-set dense segmentation, but lacking a comprehensive benchmark limits our ability to fully evaluate model performance across diverse open-set scenarios.

{
    \small
    \bibliographystyle{ieeenat_fullname}
    \bibliography{main}
}

\clearpage
\appendix
\setcounter{page}{1}
\maketitlesupplementary


\section{Experiments}
This section introduces more experiment details about \name, including dataset and task settings in Sec.~\ref{sec:supp_dataset}, the exact $3$$\times$$3$ super-patch arrangement in Sec.~\ref{sec:supp_superpatch}, and the mechanism of refinement script that aims to stimulate human behavior in Sec.~\ref{sec:supp_refine}.

\subsection{Refinement Mechanism}
\label{sec:supp_refine}

\myparagraph{Semantic Segmentation} With segmentation result $\mathbf{M}_{sem}$, we first calculate a confusion matrix between GT. 
Based on the IoU metric, which is the ratio of intersection and union, we propose Union minus Intersection (UmI):

\vspace{-10pt}
\begin{align}
\begin{aligned}
    UmI = Union - Intersection.
\end{aligned}
\end{align}

\noindent
UmI shows the wrongness of prediction for each category. We then pick the five highest UmI results for further refinement. Finally, the situation and prompt are determined by the recall rate with the following algorithm:

\begin{algorithm}
    \caption{Judge Refinement Situation}
    \scriptsize
    \KwIn{matrix, recall, cat\_idx}
    \KwOut{situation}
    \Comment{matrix (Tensor), confusion matrix, shape (N, N)}
    \Comment{recall (List[Float]), list of recall rate}
    \Comment{cat\_idx (List[Int]), list of category to refine}
    \For{i, idx in enumerate(cat\_idx)}{
        \eIf{recall[i] < 0.2}{
            \Comment{matrix[:, n], pixel belong to class n}
            \Comment{matrix[m, :], pixel predicted as class m}
            other\_cat\_iou = matrix[:-1, idx].max() / matrix[:, idx].sum()
            
            \eIf{other\_cat\_iou > 0.5}{
                \Comment{incorrect classification}
                situation = ``category"
            }{
                \Comment{missed detections}
                situation = ``missed"
            }
        }{
            \Comment{correct classification}
            situation = ``mask"
        }
    }
\end{algorithm}

\myparagraph{Instance Segmentation} With N instance prediction after processing, we first calculate an IoU matrix with M GT and keep the highest IoU result for each prediction as the matching result. Then, after descending sort by objectness score, we select ten predictions under 50 IoU for the refinement. Finally, the situation is selected from the first two situations, determined by the correctness of the classification result.

\myparagraph{Referring Segmentation} Because referring segmentation predicts mask solely, we use the first situation (mask) to refine every prediction.

\subsection{Super-patch}
\label{sec:supp_superpatch}

\begin{figure}[h]
  \centering
    \includegraphics[width=0.90\linewidth]{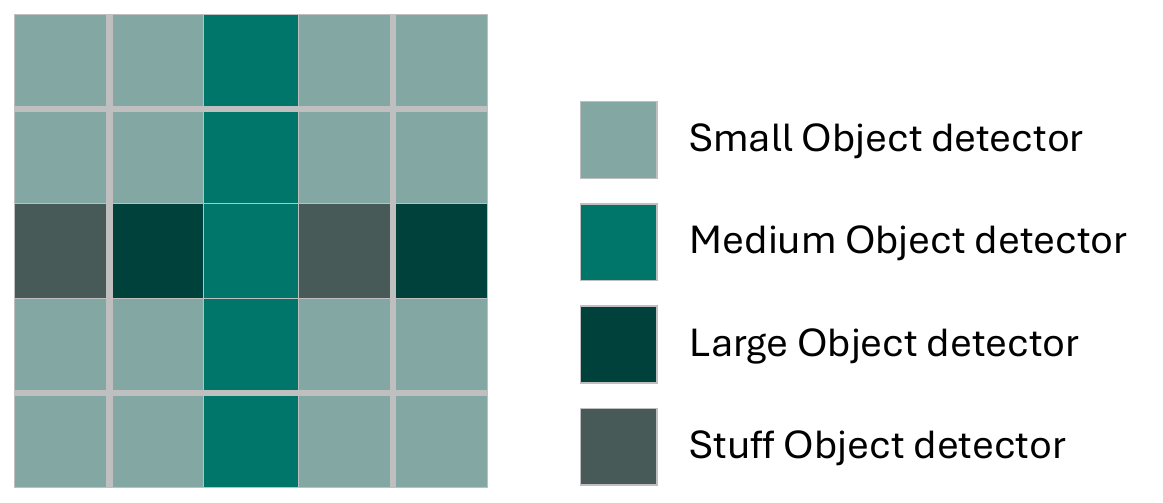}
    \caption{$3$$\times$$3$ super-patch arrangement.}
    \vspace{-5mm}
  \label{fig:supp_superpatch}
\end{figure}

As mentioned in the main paper, in the default experiment we assign 4 patches for small object detection, 3 patches for medium objects, and one each for large object and stuff region. 
In Fig.~\ref{fig:supp_superpatch} we show how exactly the patches are arranged. A thick gray line indicates the borderline between each super-patch area.

\vspace{-7pt}

\subsection{Training convergence}
\begin{figure}[h]
  \centering
    \includegraphics[width=\linewidth]{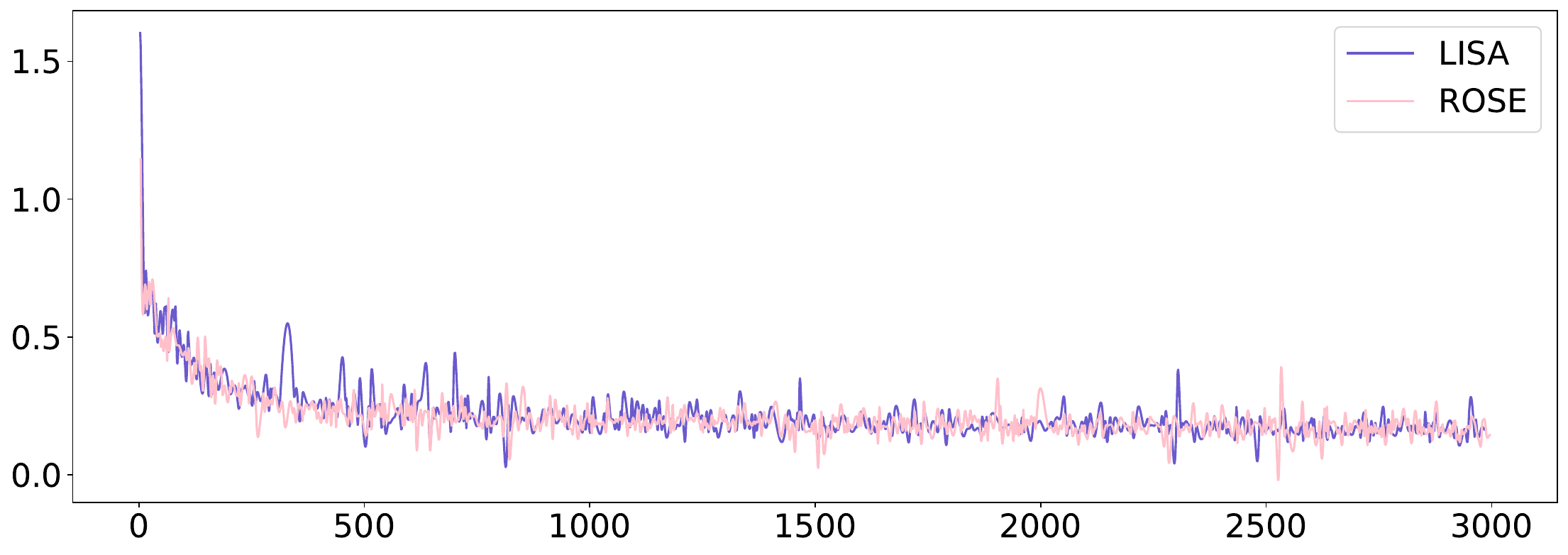}
    \caption{Mask loss during training.}
    \vspace{-5mm}
  \label{fig:supp_loss}
\end{figure}

We compared the loss convergence of ROSE with LISA. ROSE requires a bit more trainable parameters (4.8\%) than LISA (3.7\%), but it affects convergence little in the training stage according to Fig.~\ref{fig:supp_loss}.

\vspace{-7pt}

\subsection{Dataset and Task}
\label{sec:supp_dataset}
\myparagraph{Semantic Segmentation} In the training stage, we use instance-level supervision for thing categories, instead of semantic-like supervision. Because we want our model to distinguish different identities as the category may change with the granularity. And instance-level supervision is more reasonable for patch-unit prediction. 
In the inference, following Mask2former, we stack prediction within the same category and get N-channel mask $\mathbf{M}_{sem} \in \mathbb{R}^{N \times H \times W}$. Here, $N$ is the number of categories of the dataset plus one non-object channel. 
The prompts we employed look like this 
``\texttt{\textbf{User}}: \texttt{<IMAGE>} \texttt{\small \justify Can you segment this image? Please respond with category names and corresponding segment masks.}".

\begin{figure*}[hbt!]
  \centering
    \includegraphics[width=0.8\linewidth]{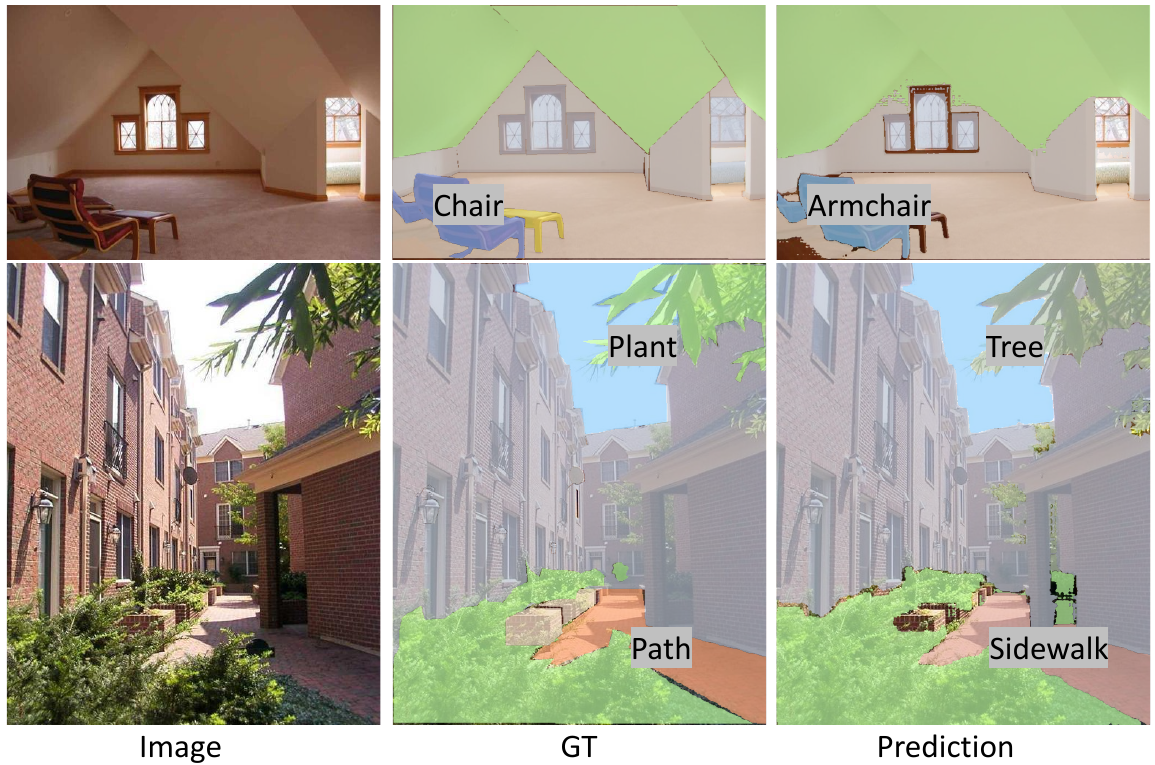}
    \caption{\textbf{Failure cases.} We show some samples of typical failure scenario in the ADE20k dataset. Wrong classification results are labeled.}
    \label{fig:failure_case}
\end{figure*}

\myparagraph{Instance Segmentation} 
In inference, we first use a threshold to filter predicted instances, and then NMS post-process is conducted to get the final results.
And the corresponding prompt like
``\texttt{\textbf{User}}: \texttt{<IMAGE>} \texttt{\small Please segment all the foreground instances in this image.}".

\myparagraph{Referring Segmentation} 
Following LISA, with the \textit{\{description\}} annotation in each dataset, one of the prompts we use is shown below
``\texttt{\textbf{User}}: \texttt{<IMAGE>} \texttt{\small What is} \texttt{\{description\}} \texttt{\small in this image? Please output the segmentation mask.}".


\myparagraph{Segmentation Refinement} 
Here, we show some examples of the prompts used in different situations: 1) the classification is correct but the segmentation mask is corrupted:
``\texttt{\textbf{User}}: \texttt{<IMAGE,MASK>} \texttt{\small This segmentation mask is incomplete, please ensure the entire object is captured.}".
2) incorrect classification with imperfect segmentation:
``\texttt{\textbf{User}}: \texttt{<IMAGE,MASK>} \texttt{\small The category of this segmentation result is wrongly predicted as \{category\}, please correctify this.}".
3) missed detections:
``\texttt{\textbf{User}}: \texttt{<IMAGE,MASK>} \texttt{\small Please segment target region with mask and corresponding category.}".

\section{Visualizations}

\subsection{Failure cases}
Fig.~\ref{fig:failure_case} shows the typical flaw caused by granularity differences (``plant" and ``tree"). Due to the inherent limitations of existing evaluation methods. Despite the correctness, it is evaluated as wrong. We think it is an important point for future research.

\subsection{Qualitative results}
\begin{figure*}[hbt!]
  \centering
    \includegraphics[width=1.0\linewidth]{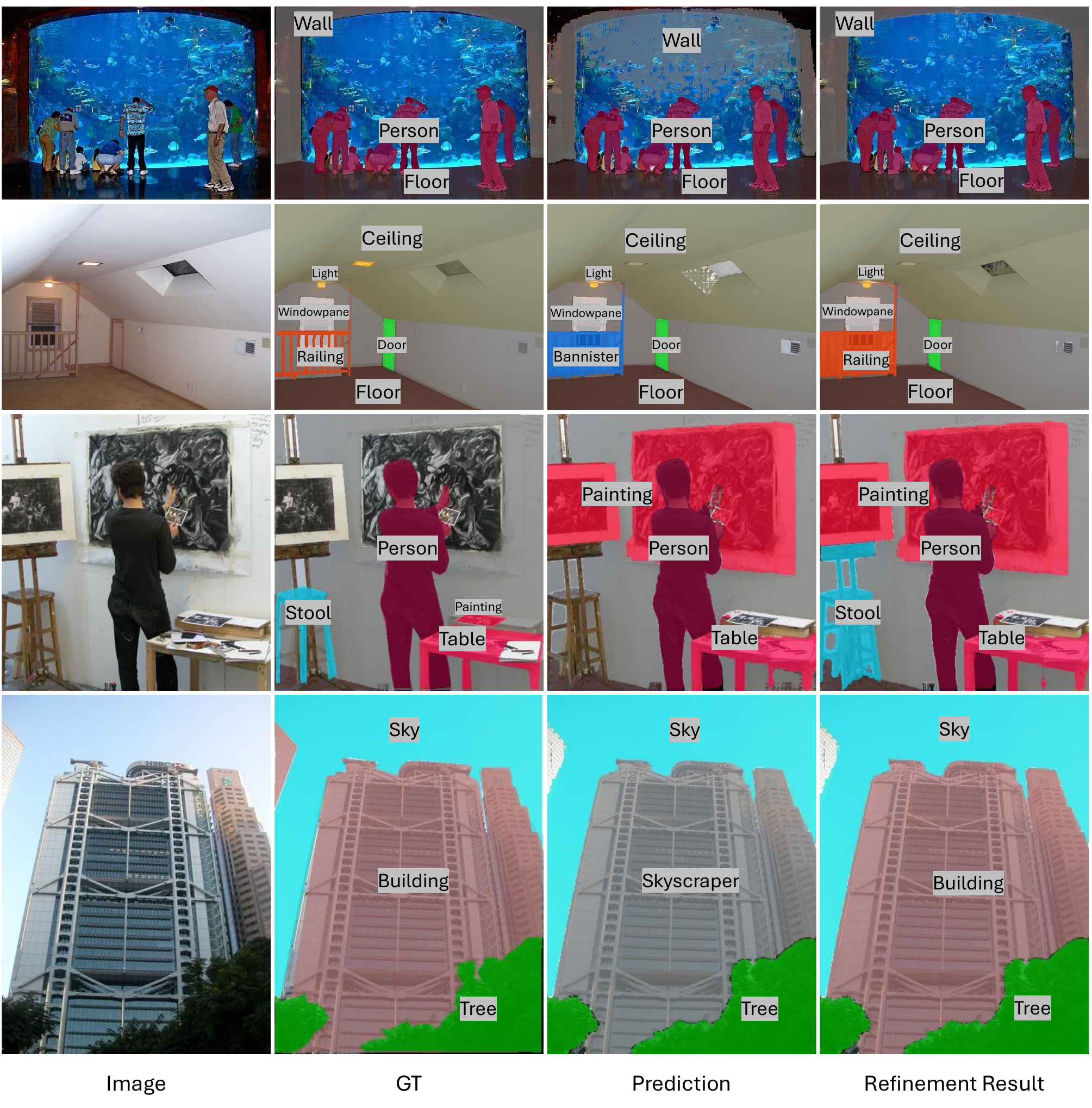}
    \caption{Qualitative Results on ADE-20k. Input image, GT, original prediction result, and result after refinement are shown. The corresponding category predictions are tagged on each prediction result.}
    \vspace{-9pt}
  \label{fig:supp_visual}
\end{figure*}

\begin{figure*}[hbt!]
  \centering
    \includegraphics[width=1.0\linewidth]{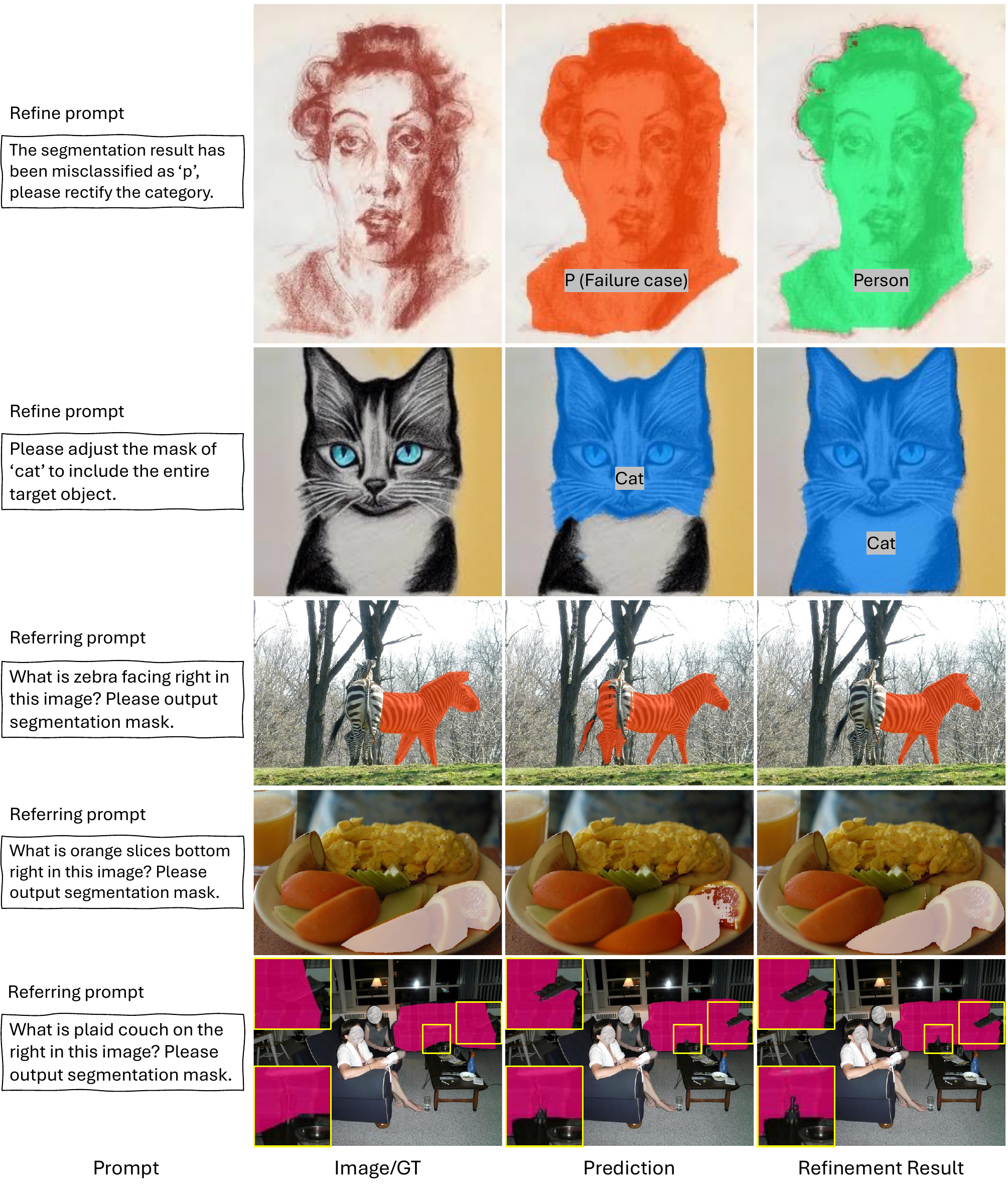}
    \caption{Qualitative Results on cross-domain images and referring tasks. Prompt, input image, GT, original prediction result and the result after refinement are shown. The corresponding category predictions are tagged on each prediction result for cross-domain samples.}
  \label{fig:supp_visual_1}
\end{figure*}

We show more qualitative results here, Fig.~\ref{fig:supp_visual} shows some results on ADE-20k, and Fig.~\ref{fig:supp_visual_1} shows results on cross-domain images and RefCOCO. We pick up various samples to cover all of the pre-defined segmentation refinement situations, which verifies the stability and effectiveness of \name\ and its refinement mechanism.

\end{document}